\documentclass[preprint,12pt]{elsarticle}

\usepackage{amssymb}
\usepackage{framed}
\usepackage{array}
\usepackage{tabularx}
\usepackage{subcaption}
\usepackage{hyperref}

\journal{Computer Vision and Image Understanding}

\begin{document}

\begin{frontmatter}



\title{Bayesian ensemble learning for image denoising}


\author[rvt]{Hyuntaek Oh\corref{cor1}}
\ead{hyuntaekoh@ku.edu}

\cortext[cor1]{Corresponding author}

\address[rvt]{University of Kansas, Lawrence, KS 66045}

\begin{abstract}
Natural images are often affected by random noise and image denoising has long been a central topic in Computer Vision. Many algorithms have been introduced to remove the noise from the natural images, such as Gaussian, Wiener filtering and wavelet thresholding. However, many of these algorithms remove the fine edges and make them blur. Recently, many promising denoising algorithms have been introduced such as Non-local Means, Fields of Experts, and BM3D. 
In this paper, we explore Bayesian method of ensemble learning for image denoising. Ensemble methods seek to combine multiple different algorithms to retain the strengths of all methods and the weaknesses of none. Bayesian ensemble models are Non-local Means and Fields of Experts, the very successful recent algorithms. The Non-local Means presumes that the image contains an extensive amount of self-similarity. The approach of the Fields of Experts model extends traditional Markov Random Field model by learning potential functions over extended pixel neighborhoods. The two models are implemented and image denoising is performed on natural images. The experimental results obtained are used to compare with the single algorithm and discuss the ensemble learning and their approaches. Comparing to the results of Non-local Means and Fields of Experts, Ensemble learning showed improvement nearly 1dB.

\end{abstract}


\end{frontmatter}


\section{Introduction}
\label{}
Some natural images contain some degree of natural and artificial noise. These noises usually affect the visual quality of the original images, so the goal of image denoising is to reconstruct a reasonable estimate of the original image from the noisy image. Ideally, the resulting denoising image will not contain any noise or added artifacts.

In the past few decades, many novel approaches have been proposed for image denoising \cite{rudin1992nonlinear, ghael1997improved, olshausen1997sparse, buades2005non, roth2009fields, dabov2006image}. One striking aspect of image denoising research is that a wide array of denoising strategies have remained popular, and in spite of vastly different approaches, many of these algorithms produce reasonably similar performance in terms of peak signal-to-noise ratio (PSNR). For example, Fields of Experts (FoE) pursues an entirely parametric approach, by training Markov random fields with large 5 x 5 cliques to capture the statistics of small image patches. Over a set of six canonical images, FoE attained a PSNR of 30.24dB for Gaussian noise with $\sigma = 20$. A Gaussian scale mixtures also uses a parametric approach, and captures the joint statistics of neighboring Gabor filter coefficients. Over the same set of six images, mean PSNR was 30.78dB for $\sigma = 20$. Sparse dictionaries is a method that seeks to identify an optimal set of image patches to form the basis of a sparse L1 norm. Over the same set of images, sparse dictionaries achieved a mean PSNR of 30.57dB when the dictionary was trained from natural images, and 31.01dB when trained on the noisy input image. Another method that exploits patterns found within the noisy input image is NL-means. However, NL-means uses a wholly non-parametric approach, by identifying similar patches within the noisy input image and averaging these together, weighed according to similarity and proximity. NL-means achieves a PSNR of 30.37dB for $\sigma = 20$ on the same set of images. BM3D is an algorithm with a similar strategy, but uses more sophisticated methods to combine similar image patches.

This brief list of algorithms includes some that are parametric and others non-parametric, some that focus on matching natural scene statistics and others that focus on utilizing patterns from within the noisy input image, and some that use generatively trained probabilistic models, some discriminatively trained probabilistic models, and others that do not use probabilistic models at all. Numerous additional differences are evident between the implementation details of each approach. In spite of these significant differences in strategy, performance is reasonably similar between these varied algorithms. Continual improvements to denoising algorithms regularly change the dominant approach, and no category of denoising strategies has produced a clear enough victor to discourage further research in any other category.

On the surface, this observation may suggest that image denoising algorithms are converging to some upper-bound on denoising performance. However, it is worth noting that each method is regarded as having different advantages and disadvantages. Strategies that perform best for low noise levels may not perform as well for high noise levels. Input images that contain many regular textures or patterns often benefit from non-local methods, while images with less internal regularities may benefit from algorithms trained from large suites of natural images. Additionally, one important quality of denoising methods is the ability to preserve sharp edges while removing noise. Methods that fail in this regard produce output that appear over-blurred. Non-local methods often perform well at maintaining sharp edges, as demonstrated by their residual images (the denoised image minus the true image). These residual images show that methods like NL-means perform similarly near edges as they do near smooth regions. Other methods such as FoE show higher residual error near edges. Since FoE achieves a similar overall PSNR, it suggests that performance within smooth regions is higher for FoE.

When multiple regression algorithms produce similar performance using significantly different approaches, and with distinct advantages and disadvantages, those algorithms are highly suitable for combination using ensemble learning methods. Ensemble learning is a method of combining multiple (possibly weak) predictors to produce one unified predictor of greater accuracy. While ensemble learning is a common and successful technique in machine learning, it has not been applied to image denoising. Ensemble learning methods benefit when the constituent algorithms are significantly different from one another. In this paper, we apply Bayesian ensemble learning methods to combine two of the most distinct denoising methods: Fields of Experts and NL-means. We use 40 natural images from the Berkeley Segmentation Benchmark for training \cite{martin2001database}. Another set of 80 natural images from the Berkeley database are used for testing, along with 6 canonical images such as Barbara and Lena. For each level of input noise, the ensemble method achieved statistically significant improvement over both FoE and NL-means.

\section{Background}
\subsection{Image Denoising}
The goal of image denoising is to reconstruct the original image from the noisy image,
\begin{equation}
y(i) = x(i) + n(i)
\end{equation}•
where $y(i)$ is the observed image, $x(i)$ is the original image; and $n(i)$ is the noise value at pixel $i$. Gaussian white noise is widely used in natural images for image denoising. The noise value, $n(i)$, is the Gaussian white noise values with known variance $\sigma^2$ and zero mean. The Gaussian white noise models are made by adding random values to the original images. The ideal denoising algorithm is to remove the noise, $n(i)$, and recover the original image, $x(i)$.

Previous methods such as Gaussian or Wiener filtering attempt to separate the image into the two parts, the smooth and oscillatory part, by removing the high frequency from the low frequency \cite{ghael1997improved, portilla2003image}. This would result in a loss of fine edges in the denoised image. Low frequency noise will remain in the image even after denoising. Therefore, new algorithms have been introduced recently such as Non-local means \cite{buades2005non} and Fields of Experts \cite{roth2009fields}.

\subsection{Fields of Experts}
\begin{figure}[h]
\centering
\includegraphics{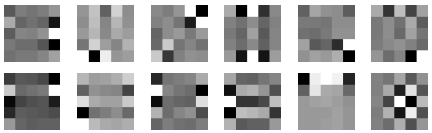}
\caption{Selection of the 5x5 filters learned from the training data}
\label{fig:filter}
\end{figure}

Fields of Experts (FoE) was proposed by Stefan Roth and Michael J. Black \cite{roth2005fields}. The goal of the FoE is to develop a framework for learning rich, generic prior models of natural images. To learn potential functions through extended neighboring pixels, a Markov Random Field model was used in the FoE. The key in the FoE is to extend Markov Random Field by modeling the local field potentials with learned filters \cite{roth2005fields}. To do this, Products of Experts were used \cite{roth2009fields}. In comparison with prior Markov Random Field approaches, all parameters in the FoE model are learned from a set of training data \cite{roth2005fields}. One of the parameters is the clique size and 5x5 was used. Those models prior probability of images can be calculated with the following formula:
\begin{equation}
P(\vec{I}) \propto \prod_C \prod_{i=1}^K \left(1+\frac{1}{2}\left(\vec{I_c}\cdot \vec{J_c}\right)^2 \right) ^{-\alpha_i}
\label{eq:foe}
\end{equation}•
where $I_c$ is 5x5 image patch, and filter $J_c$ represents especially unlikely image patches obtained by training the FoE model on an general image database. 20,000 image patches were selected randomly from the Berkeley Segmentation database, and the image patches are used for the training data \cite{martin2001database}. Figure \ref{fig:filter} shows the unlikely 5x5 image patches which were used as the filter, $J_c$, in the formula \ref{eq:foe}.

•

Inference: For the denoising problem, the goal is to infer the most likely correction for the image given the prior and the noisy image. Given a noisy image N, we can find the denoised image D that maximizes the prior probability:
\begin{equation}
p(D|N) \propto p(N|D)p(D)
\end{equation}•

We can write the $P(N|D)$ as:
\begin{equation}
p(N|D) \propto \prod_j \exp \left(-\frac{1}{2\sigma^2} \left(D_j - N_j\right)^2  \right)
\end{equation}•
where $\sigma$ is known standard deviation, and $D_j$ and $N_j$ are the denoised and noisy image at pixel $j$, respectively. In this study, we use the FoE algorithm MATLAB code provided by Roth and Black, and use the similar parameters to get the same results of the paper.

\subsection{Non-local Means}
Non-local Means (NLM) image denoising algorithm was suggested by Antoni Buades, Bartomeu Coll, and Jean-Michael Morel. NLM presumes that the amount of pixel weighting is based on the similarity of their neighborhoods with the neighborhood of each pixel \cite{buades2005non}. Efros and Leung originally suggested the concept of the self-similarity for texture synthesis \cite{efros1999texture}.

\begin{figure}[h]
\centering
\includegraphics[width = \textwidth]{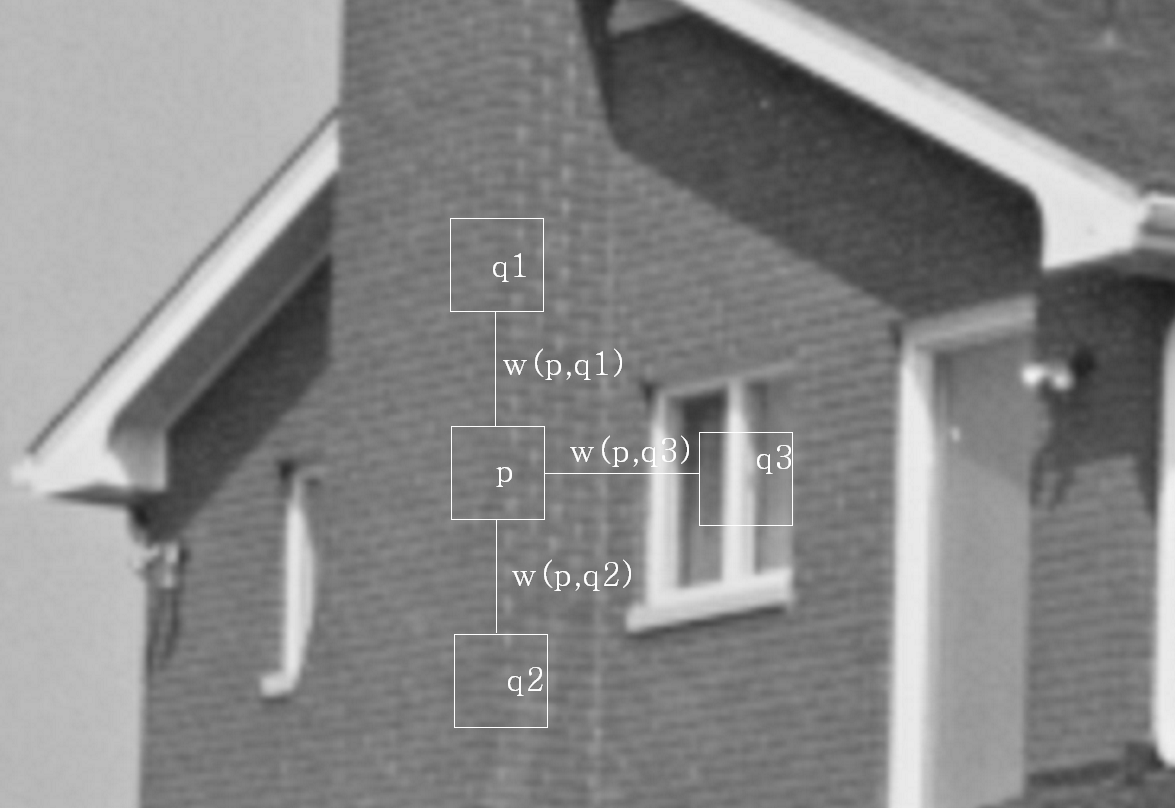}
\caption{Scheme of NLM strategy \cite{buades2005non}}
\label{fig:nlm}
\end{figure}•

Figure \ref{fig:nlm} shows the scheme of NLM strategy. The figure shows four different pixels: p, q1, q2, and q3. Similar neighborhoods to p’s neighborhood could be found in most pixels in the same column of p such as q1 and q2. Similar pixel neighborhoods give a large weight, w(p, q1) and w(p, q2), while much different neighborhoods give a small weight w(p, q3).

To compute each pixel i of the NLM denoising image, the following formula was used \cite{buades2005non}
\begin{equation}
NL[v](i) = \sum_{j \in I} w(i,j)v(j)
\end{equation}•
where $v$ is the noisy image and $v = \{v(i)|i \in I\}$, and weights $w(i,j)$, which rely on the similarity between pixel $i$ and $j$, meet the following conditions $0 \leq w(i,j) \leq 1$ and $\sum_{j \in I} w(i,j) = 1$. A neighborhood should be defined to compute the similarity. The weighted Euclidean distance is used to measure the similarity, and the following formula is implemented to calculate the Euclidean distance of noisy neighborhoods:
\begin{equation}
E\|v(N_i) - v(N_j)\|_{2,a}^2 = \|u(N_i) - u(N_j)\|_{2,a}^2 + 2\sigma^2
\end{equation}•
where $v(N_i)$ and $v(N_j)$ are the gray scale vectors, $N_i$ and $N_j$ are the square neighborhood of fixed size and centered around a pixel $i$ and $j$, respectively; and $a$ is the Gaussian kernel’s standard deviation. The weights $w(i,j)$ can be calculated with the following formula:
\begin{equation}
w(i,j) = \frac{1}{Z(i)} e^{-\frac{\|v(N_i) - v(N_j)\|_{2,a}^2}{h^2}}
\end{equation}•
where $Z(i)$ is the normalizing constant
\begin{equation}
Z(i) = \sum_j e^{-\frac{\|v(N_i) - v(N_j)\|_{2,a}^2}{h^2}}
\end{equation}•
and the parameter $h$ satisfies as a filtering degree.

\section{Application of Ensemble learning}
As discussed in the introduction, in spite of relatively comparable performance, image denoising techniques like FoE and NLM use very different methodologies and underlying philosophies for image denoising. This would seem to make these two algorithms strong candidates for ensemble learning methods which may be able to produce a single denoising algorithm that retains the advantages of both FoE and NLM. However, this problem differs from traditional ensemble learning problems in several important ways, which result in both advantages and disadvantages in comparison with standard ensemble learning scenarios.

One significant difference is that ensemble learning is typically used to combine multiple models of the data or the predicted output, where each constituent algorithm was trained on labeled data to optimize the parameters of the algorithm. In contrast, many image denoising methods utilize no training at all. Methods like NLM are derived primarily from a theoretical analysis of the image denoising problem, rather than machine learning techniques. NLM does have some parameters, such as the size of the search window, but these parameters are not expected to affect the behavior of the algorithm significantly, and so training is not emphasized.

The lack of training in the constituent algorithms makes image denoising ineligible for some of the most powerful aspects of ensemble learning. Many ensemble learning methods, such as boosting \cite{schapire1990strength} or bootstrap aggregation \cite{efron1993introduction}, achieve higher performance by training each constituent algorithm on different subsets of available training data. In that way, one algorithm becomes an “expert” on one type of input, while another algorithm specializes in another type of input. Because NLM does not rely on training or make use of labeled data, such techniques are not possible here.

Another important difference is that the bulk of ensemble learning methods are designed for classification problems, where the predicted output is either binary, or at least discrete. For classification problems, the outputs of each constituent algorithm are typically combined using a voting scheme: whichever class receives the most votes (possibly weighted by the performance of the constituent algorithm) is the output of the ensemble learning algorithm. Ensemble learning for continuous-valued output (known as regression) is not entirely uncommon \cite{breiman1996stacked}. In these cases, outputs of the constituent algorithms are usually combined by averaging. For image denoising, averaging can improve error metrics based on sum squared error. Unfortunately, though, averaging multiple denoised images may lead to blurry edges and other undesirable artifacts that reduce perceived denoising quality. Each denoising algorithm must hypothesize the most probable image structure that was obscured by noise in the input image. When these hypotheses disagree, averaging them together can result in two faint but duplicate differing image structures, rather than a single consistent denoised image. For example, consider the case where noise has obscured the precise location of an edge between two regions. If NLM hypothesizes an edge in one location, and FoE hypothesizes an edge in a different location, an averaging procedure will produce two faint edges, which is generally an unlikely result. We would prefer an ensemble technique that could select the single edge location that appears most likely. Ideally, an ensemble method should be capable of selecting a single coherent denoising hypothesis rather than combining multiple outputs naively. This advantage of traditional ensemble methods may provide one explanation as to why ensemble methods have not been applied to image denoising in the past.

If a large number of constituent algorithms are available, more sophisticated schemes for combining algorithm output might be possible, where outlier responses were given less weight. Such techniques might avoid the pitfalls of averaging methods listed above. Unfortunately, only a handful of competitive image denoising algorithms exist.

Finally, a third major difference is that some constituent denoising algorithms (in particular, FoE) provide not only a single hypothesis denoised image, but also provide a probability distribution over the space of all possible denoised images. Recall that FoE outputs not only a denoised image $D_{FoE}$, but also provides a probability distribution over denoised images $P_{FoE}(D|N)$, where $N$ is the input noisy image. This distribution allows us to quantify the uncertainty in the FoE solution, or potentially, to measure the likelihood of the outputs of other denoising algorithms within the FoE model. As we describe below, the availability of a probabilistic model of the output space is a great advantage for applying ensemble method to the image denoising problem, because it provides a possible solution to the disadvantages of averaging described above.

If all constituent algorithms provide a distribution over the space of hypothesis, then Bayesian ensemble learning methods can be used to combine each distribution into a single distribution. In particular, the choice of which model is superior for a particular input can be treated as a latent variable. Specifically, suppose we wish to predict output y given input $x$, and we have a database of labeled training examples $Z$. Also, suppose we have $M$ different probabilistic predictors which each provide a distribution $P_m(y|x)$ over possible outputs. Then we can write the probability of possible outputs y as:
\begin{equation}
P(y|x,Z) = \sum_{m=1}^M P(y,m|x,Z) 
\end{equation}•
\begin{equation}
P(y|x,Z) =  \sum_{m=1}^M P(y|m,x,Z) P(m|x,Z) = \sum_{m=1}^M P(y|x,Z) P(m|Z)
\end{equation}•

Thus, the Bayesian ensemble distribution is simply the weighted sum of each constituent distribution, weighted by performance over the training data $Z$. The downside of Bayesian ensemble methods is that the ensemble distribution $\sum_{m=1}^M P(y|x,Z) P(m|Z)$ is often computationally demanding to optimize. However, the advantage is that outputs y that score highly in the ensemble distribution must be considered probable according to all of the constituent algorithms, especially those that were most successful on the training data. If Bayesian ensemble methods were applied to the image denoising problem, this would be eliminate the pitfall of averaging two denoising outputs together. The mean of two plausible solutions may not itself be plausible, but the optimum of a Bayesian ensemble distribution largely satisfied both models simultaneously.

Unfortunately, Bayesian ensemble learning methods cannot be applied in the denoising problem, because many constituent denoising algorithms do not provide a probability distribution over denoised images. For example, NLM, we must combine a mixture of probabilistic and non-probabilistic models. To our knowledge, this circumstance has not been studied explicitly in past ensemble learning methods. Our goal is to retain the advantage of purely Bayesian ensemble approaches: the ensemble method should produce an output that internally consistent, considered highly plausible by the FoE probability distribution, while simultaneously resembles the NLM output. Additionally, we need to find a method that is efficient. Bayesian ensembles are often computationally intensive to optimize, in part because, being summations, they do not factorize.

Our approach is to treat the NLM output as a known, given quantity. Thus, we want to model $P(I|N,D_{NL})$, where $I$ is the hypothetical original image, $N$ is the noisy image, and $D_{NL}$ is the output of NLM. By applying Bayes rule, we can write
\begin{equation}
P(I|N,D_{NL}) = P(N|I,D_{NL})P(I|D_{NL})/P(N|D_{NL}) = \frac{P(N|I)P(D_{NL}|I)P(I)}{P(N|D_{NL})P(D_{NL})}
\label{eq:11}
\end{equation}•
\begin{equation}
P(I|N,D_{NL}) \propto P(N|I)P(D_{NL}|I)P(I)
\label{eq:12}
\end{equation}•
where terms that do not depend on $I$ can be ignored as constants. The term $P(N|I)$ is simply the noise model, which is a Gaussian of mean $I$ and variance given by the strength of the image noise. The prior over noiseless images, $P(I)$, can be taken from the FoE model, given by equation \ref{eq:foe}.

To complete the ensemble model, we must choose a model for $P(D_{NL}|I)$. One known strength of the NLM method is that the image residual, defined as $I-D_{NL}$, shows little image structure \cite{buades2005non}. In other words, edges and features that are visible in the noisy image are faint or not visible in the residual. In comparison, other leading denoising images often produce residuals that retain structure from the noisy image. This advantage of NLM is believed to stem from the methodology used by NLM. Trained methods like FoE base their models for image structure such as edges entirely from databases of natural images. In contrast, NLM acquires statistics of image structure from the noisy input image itself. For example, FoE may misjudge the spatial scale, or sharpness, of edges if evidence of the edge is weak within the image, and other false spatial of edges solely by comparing against similar edges within the noisy image, and the edges with less common scales are less likely to be biased. For these reasons, strong structure visible in the noisy input image is less likely to be visible in the residuals of NLM outputs.

One consequence of this observation is that the residual of NLM can be approximated as white noise. In particular, the residual of NLM is known to closely resemble the statistical structure of the additive noise. This allows us to define $P(D_{NL}|I)$ accordingly. In the case of input images with additive Gaussian noise, $P(D_{NL}|I)$ is Gaussian centered at $I$, with some $\sigma_{NL} < \sigma$:
\begin{equation}
p(D_{NL}|I) \propto \exp \left(-\frac{\sum_{x,y}\left(I(x,y) - D_{NL}(x,y) \right)^2}{2\sigma_{DL}^2} \right)
\label{eq:13}
\end{equation}•

This completes the definition of our ensemble model. Note that this model completes our objective: images that optimize the ensemble distribution are simultaneously

\begin{enumerate}[1.]
\item plausible according to the Fields of Experts model
\item similar to the Non-local Means output
\item close to the noisy input image
\end{enumerate}

We also must ensure that our model can be optimized efficiently. Here, we observe that our model can be simplified into a form that is very similar to the FoE probabilistic model \cite{roth2009fields}. We can start with the formula \ref{eq:12}.
\begin{equation}
p(N|I)p(D_{NL}|I) \propto \exp \left(\frac{(N-I)^2}{2\sigma_N^2} + \frac{(D_{NL} - I)^2}{2\sigma_{NL}^2} \right) \propto \exp \left(\frac{(N_{pseudo} - D)^2}{2\sigma_{pseudo}^2} \right)
\label{eq:14}
\end{equation}•

In formula \ref{eq:13}, the left side shows a component of the ensemble learning model, and the right side shows the same model re-arranged to have the same structure as FoE \cite{roth2009fields}. This new form introduces new variables such as $N_{pseudo}$ and $\sigma_{pseudo}$. We can get these parameters $N_{pseudo}$ and $\sigma_{pseudo}$ by comparing the left and right side. In the left side, the parameters such as $N$, and $D_{NL}$ are all functions.

\begin{equation}
\frac{(N^2-2NI+I^2)}{2\sigma_N^2} + \frac{(D_{NL}^2 - 2D_{NL}I+I^2)}{2\sigma_{NL}^2} \propto \frac{(N_{pseudo}^2 - 2N_{pseudo}I+I^2)}{2\sigma_{pseudo}^2}
\label{eq:15}
\end{equation}•

Therefore, the parameter $\sigma_{pseudo}$ and $N_{pseudo}$ are
\begin{equation}
\sigma_{pseudo}^2 = 1/\{(1/\sigma_N^2)+(1/\sigma_{NL}^2)\}, N_{pseudo} = N\cdot \alpha + D_{NL} \cdot \beta
\label{eq:sigma_pseudo}
\label{eq:16}
\end{equation}•
where
\begin{equation}
\alpha = (1/\sigma_N^2)/\{(1/\sigma_N^2)+(1/\sigma_{NL}^2) \}, \beta = (1/\sigma_{NL}^2)/\{(1/\sigma_N^2)+(1/\sigma_{NL}^2) \}
\label{eq:17}
\end{equation}•

Because the ensemble learning is usually computationally intensive, choosing the method of combining two algorithms, such an NLM and FoE in this case, allows us to do with extra computational effort and extra coding to build the ensemble learning.

In theory, $\sigma_{NL}$ could be estimated empirically by finding the standard deviation of the residuals of NLM outputs on natural images for each noise level σ. In practice, we found that superior performance was attained by learning $\sigma_{NL}$ from a set of training images.\\

\begin{figure}[h]
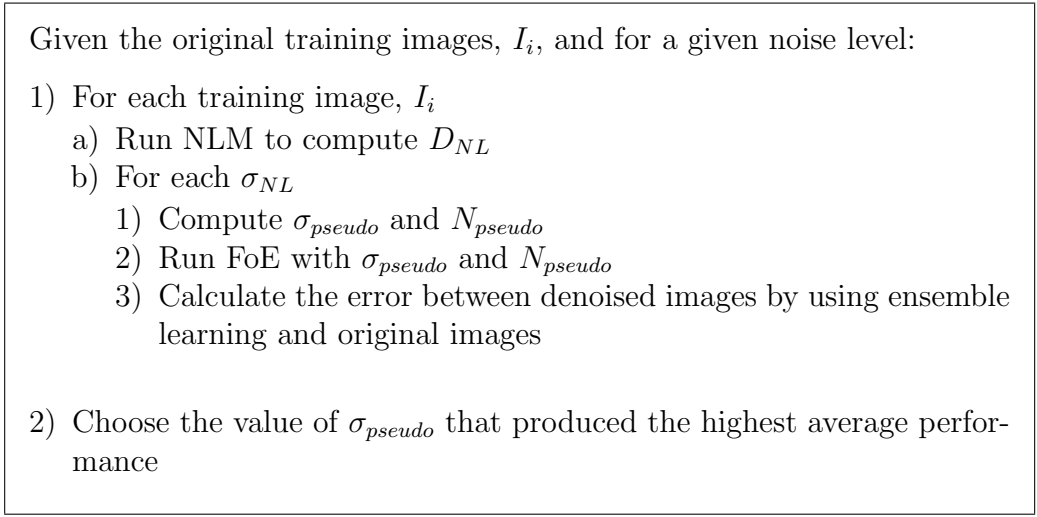

\begin{framed}
Given the original training images, $I_i$, and for a given noise level:
\begin{enumerate}[1)]
\item For each training image, $I_i$
\begin{enumerate}[a)]
\item Run NLM to compute $D_{NL}$
\item For each $\sigma_{NL}$
\begin{enumerate}[1)]
\item Compute $\sigma_{pseudo}$ and $N_{pseudo}$
\item Run FoE with $\sigma_{pseudo}$ and $N_{pseudo}$
\item Calculate the error between denoised images by using ensemble learning and original images
\end{enumerate}•
\end{enumerate}•
\item Choose the value of $\sigma_{pseudo}$ that produced the highest average performance
\end{enumerate}
\end{framed}
\caption{Pseudo code for training $\sigma_{pseudo}$}
\label{fig:pseudo_code}
\end{figure}

Figure \ref{fig:pseudo_code} shows the pseudo code for training $\sigma_{pseudo}$. First, a noisy image could be made with the original image by using the Gaussian distribution. The same noisy images were used for NLM and FoE in this model. The parameters $\sigma_{pseudo}$ and $N_{pseudo}$ were calculated with the denoised image, $D_{NL}$, which was done by the NLM, $\sigma_N$, the original input Gaussian sigma value and $\sigma_{NL}$, another sigma value from the NLM denoised image. The ensemble learning algorithm was adapted to the NLM denoised image, $D_{NL}$, with the learned parameters, $\sigma_{pseudo}$ and $N_{pseudo}$. The PSNR could be computed by using the original images and denoised images after performing the ensemble learning. Same process was executed with different number of the input Gaussian noise values, $\sigma_N$. 

For natural image denoising, 40 images from Berkeley Segmentation Benchmark were used for training $\sigma_{pseudo}$. The selection of $\sigma_{pseudo}$ is very important in our proposed ensemble learning. Based on the best selection of $\sigma_{pseudo}$, we used 80 natural images from Berkeley database and 6 canonical images for testing.

\section{Experimentation and Results}
\subsection{Training dataset}
First, we set aside 40 natural images randomly selected from the Berkeley Segmentation Benchmark to use as training images \cite{martin2001database}. From these images, we learned the best value of $\sigma_{pseudo}$ to use for each input noise level $\sigma$. These 40 images contain a diverse array of subject matter including architecture, landscape, people, flowers, etc.

The noisy image was obtained from the original image by adding Gaussian noise with different noise amplitude:  $\sigma = 10, 15, 20, 30, 40, 50, 75$ and 100. The NLM algorithm was used to get $D_{NL}$, and several values of $\sigma_{NL}$ for the NLM denoised image were tested. Specifically, for each training image ensemble learning was repeated with $\sigma_{NL} = 1,2,3,5,10,20,30,40,50,100,250$, and 500. We used all these sigma values, NLM denoised images and FoE algorithm to get the ensemble learning denoised images. The 5x5 filter of Fields of Experts was used to obtain the denoised images. We used 5,000 iterations to implement FoE \cite{roth2009fields}.

PSNR was calculated with for each denoised image, and the average PSNR value was computed for each combination of $\sigma$ and $\sigma_{NL}$. The average PSNR of ensemble learning with a small $\sigma_{NL}$ value, such as 1, showed a similar result with the NLM algorithm because of $N_{pseudo}$ from the formula \ref{eq:sigma_pseudo}. When the $\sigma_{NL}$ is close to a very small number, $\beta$ component in the $N_{pseudo}$ is much bigger than the $\alpha$ component, so the $N_{pseudo}$ is most likely close to the $D_{NL}$, which is the denoised image from the NLM algorithm. Likewise, when the $\sigma_{NL}$ has a large value such as 500, the $N_{pseudo}$ is very similar to the $N$, which is the original noisy image, and the denoised result of this $N_{pseudo}$ is almost the same as the FoE algorithm.

The ensemble learning achieves peak performance when the $\sigma_{NL}$ = 30,40, 50,100,100,3,5,10,40 for $\sigma$ = 10,15,20,25,30,40,50,75,100, respectively. $\sigma_{pseudo}$ and $N_{pseudo}$ can be calculated with the selected $\sigma_{NL}$ and $\sigma$ by using the formula \ref{eq:sigma_pseudo}. For example, when the $\sigma$ is 25, $\sigma_{pseudo}, \alpha$, and $\beta$ have 24.25,  0.94 and 0.06, respectively. $N_{pseudo}$ can be obtained with the calculated $\alpha$ and $\beta$.

\subsection{Results}
The performance of the ensemble denoising algorithm was evaluated over two image datasets, each separate from the training set. Performance was measured by the Peak Signal-to-noise ratio (PSNR) and the structural similarity index (SSIM) \cite{wang2004image}. PSNR is measured as

\begin{equation}
PSNR = 20 \log_{10} \displaystyle \frac{255}{\sqrt{MSE}}
\end{equation}•
where MSE denotes the mean squared error between the denoised image $N$ and the original image $I$. We also measure denoising quality using structural similarity index (SSIM). SSIM is a measure of image similarity intended to resolve certain known limitations of PSNR that may cause PSNR to disagree with human perception of image quality. In particular, a corrupted image can be perceived to be highly similar to the original if the corruptions primarily affect local contrast or brightness in certain regions. The PSNR of such a corrupted image may be very low; far lower than PSNR values produced by image corruptions that may be perceived as more severe. SSIM addresses this by normalizing for contrast and brightness both globally across the image, as well as locally, within 8 x 8 image windows. SSIM has been shown to adhere significantly closer to subjective image quality grades assigned by pools of human subjects.

\begin{figure}[ht]
        \centering
        \begin{subfigure}[b]{0.22\textwidth}
                \centering
                \includegraphics[width=\textwidth]{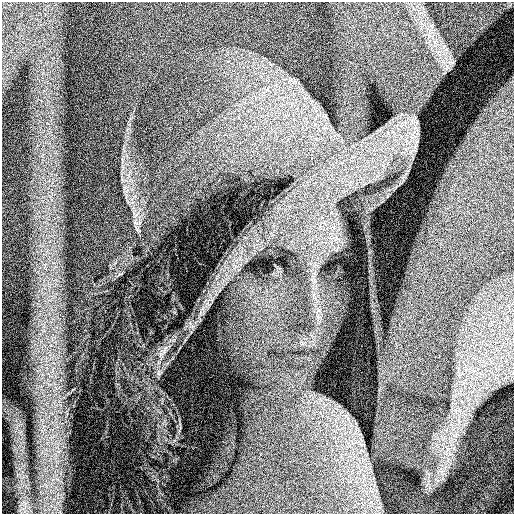}
                \caption{}
                \label{fig:noisy}
        \end{subfigure}%
        \quad 
        \begin{subfigure}[b]{0.22\textwidth}
                \centering
                \includegraphics[width=\textwidth]{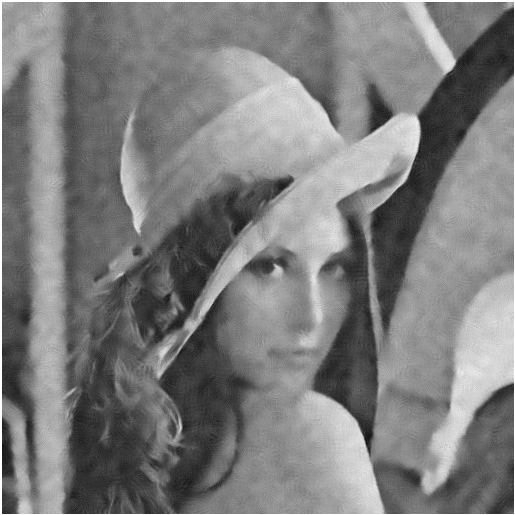}
                \caption{}
                \label{fig:NLM}
        \end{subfigure}
        ~ 
        \begin{subfigure}[b]{0.22\textwidth}
                \centering
                \includegraphics[width=\textwidth]{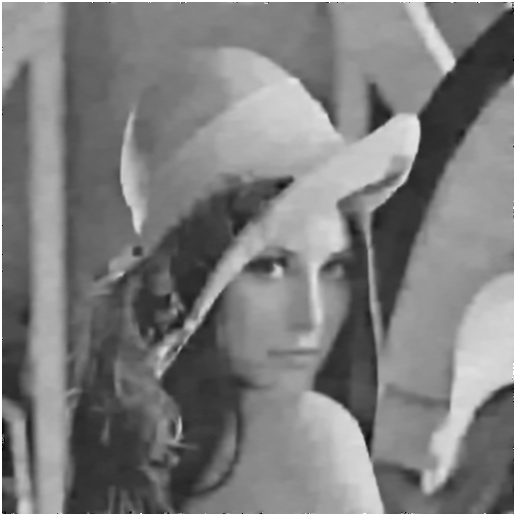}
                \caption{}
                \label{fig:FoE}
        \end{subfigure}
	~ 
        \begin{subfigure}[b]{0.22\textwidth}
                \centering
                \includegraphics[width=\textwidth]{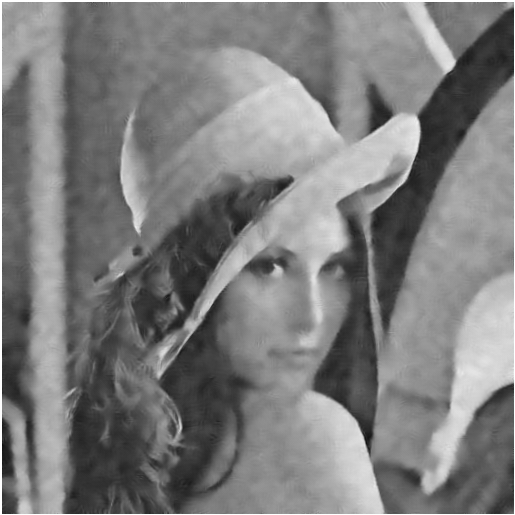}
                \caption{}
                \label{fig:ensem}
        \end{subfigure}
        \caption{Denoising Results. (a) Image with Gaussian noise, $\sigma = 50$ (PSNR = 14.18dB), (b) Denoised image using the NLM (PSNR =27.14dB), (c) Denoised image using the FoE (PSNR = 26.92dB), (d) Denoised image using the ensemble learning (PSNR = 27.62dB) with $\sigma_{NL} = 5$}\label{fig:denoising_results}
\end{figure}

The first result was done with the 6 canonical images (Barbara, boat, etc). Table \ref{table:psnr} shows the PSNR values for 6 natural images denoised by using the ensemble learning under different levels of input noise. For most images and noise levels, the ensemble learning showed an improvement in image denoising comparing to the NLM and FoE. The ensemble learning showed an especially strong improvement when the input noise value increased. Figure \ref{fig:denoising_results} shows some results of denoising images by using the NLM, FoE and ensemble learning. Comparing to these results, the ensemble learning outperformed both NLM and FoE quantitatively. With selected $\sigma_{NL}$, the PSNR result of the ensemble learning was 27.62dB. The PSNR results of NLM and FoE were 27.14dB and 26.92dB, respectively.

\begin{table}[ht]
\caption{The PSNR (dB) values for natural images denoised with the Ensemble learning  (From the left: Non-local Means, Fields of Experts and Ensemble learning)} 
\footnotesize
\centering 
\begin{tabular}{c | c c c | c c c | c c c} 

\hline 
 & \multicolumn{3}{c|}{Lena} & \multicolumn{3}{c|}{Barbara} & \multicolumn{3}{c}{Boat} \\ 
\hline
$\sigma$ & NLM & FoE & Ensem. & NLM & FoE & Ensem. & NLM & FoE & Ensem. \\
\hline
10 & \underbar{\textbf{35.17}} & 35.11 & 35.16 & \underbar{\textbf{33.72}} & 32.93 & 33.19 & 32.78 & 33.27 & \underbar{\textbf{33.27}} \\

15 & 33.36 & 33.32 & \underbar{\textbf{33.41}} & \underbar{\textbf{31.80}} & 30.25 & 30.6 & 30.97 & 31.40& \underbar{\textbf{31.42}} \\

20 & 31.97 & 32.03 & \underbar{\textbf{32.11}} & \underbar{\textbf{30.19}} & 28.41 & 28.77 & 29.54 & \underbar{\textbf{30.05}} & 30.04 \\

30 & 29.84 & 29.88 & \underbar{\textbf{29.95}} & \underbar{\textbf{27.72}} & 25.88 & 26.03 & 27.56 & \underbar{\textbf{27.96}} & 27.95 \\

40 & 28.26 & 28.13 & \underbar{\textbf{28.58}} & 25.96 & 24.17 & \underbar{\textbf{26.04}} & 26.11 & 26.18 & \underbar{\textbf{26.27}} \\

50 & 27.14 & 26.92 & \underbar{\textbf{27.62}} & 24.70 & 23.13 & \underbar{\textbf{24.78}} & 25.04 & 24.93 & \underbar{\textbf{25.28}} \\

75 & 25.03 & 24.94 & \underbar{\textbf{25.88}} & 22.91 & 22.00 & \underbar{\textbf{22.92}} & 23.27 & 23.13 & \underbar{\textbf{23.37}} \\

100 & 23.46 & 21.05 & \underbar{\textbf{24.8}} & 21.70 & 18.97 & \underbar{\textbf{21.86}} & 22.04 & 20.44 & \underbar{\textbf{22.76}} \\

\hline \hline
 & \multicolumn{3}{c|}{House} & \multicolumn{3}{c|}{Peppers} & \multicolumn{3}{c}{Fingerprint} \\ 
\hline
$\sigma$ & NLM & FoE & Ensem. & NLM & FoE & Ensem. & NLM & FoE & Ensem. \\
\hline
10 & \underbar{\textbf{35.47}} & 35.22 & 35.27 & 33.38 & 34.16 & \underbar{\textbf{34.21}} & 31.03 & 32.11 & \underbar{\textbf{32.17}}\\

15 & \underbar{\textbf{33.92}} & 33.62 & 33.74 & 31.74 & 32.05 & \underbar{\textbf{32.20}} & 29.08 & 29.64 & \underbar{\textbf{29.78}} \\

20 & \underbar{\textbf{32.61}} & 32.34 & 32.50 & 30.49 & 30.57 & \underbar{\textbf{30.76}} & 27.44 & 28.03 & \underbar{\textbf{28.16}} \\

30 & 30.05 & 30.34 & \underbar{\textbf{30.40}} & 28.11 & 28.10 & \underbar{\textbf{28.20}} & 25.09 & 25.77 & \underbar{\textbf{25.79}} \\

40 & 28.23 & \underbar{\textbf{28.74}} & 28.57 & 26.64 & 26.47 & \underbar{\textbf{26.83}} & 23.22 & \underbar{\textbf{23.58}} & 23.28 \\

50 & 26.71 & \underbar{\textbf{27.28}} & 27.17 & 25.16 & 24.95 & \underbar{\textbf{25.43}} & 21.80 & 21.47 & \underbar{\textbf{21.88}} \\

75 & 24.34 & 24.68 & \underbar{\textbf{25.15}} & 22.91 & 22.55 & \underbar{\textbf{23.35}} & 19.49 & 18.23 & \underbar{\textbf{19.54}} \\

100 & 22.85 & 19.75 & \underbar{\textbf{24.03}} & 21.5 & 18.84 & \underbar{\textbf{22.04}} & \underbar{\textbf{18.20}} & 17.63 & 17.57 \\

\hline
\end{tabular}
\label{table:psnr} 
\end{table}

\begin{figure}[ht]
        \centering
        \begin{subfigure}[b]{0.3\textwidth}
                \centering
                \includegraphics[width=\textwidth]{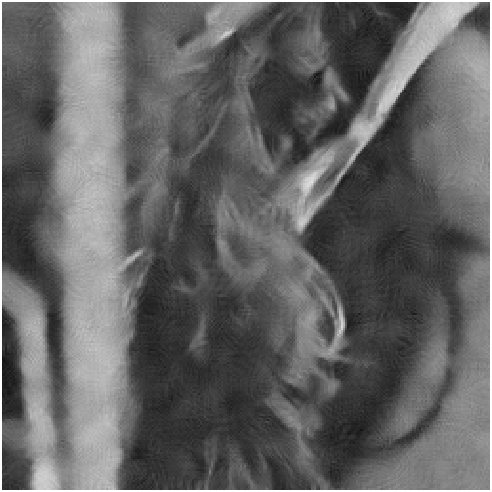}
        \end{subfigure}%
        \quad
        \begin{subfigure}[b]{0.3\textwidth}
                \centering
                \includegraphics[width=\textwidth]{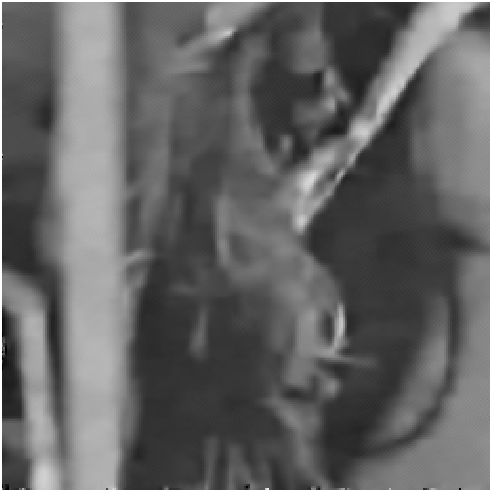}
        \end{subfigure}
        ~ 
        \begin{subfigure}[b]{0.3\textwidth}
                \centering
                \includegraphics[width=\textwidth]{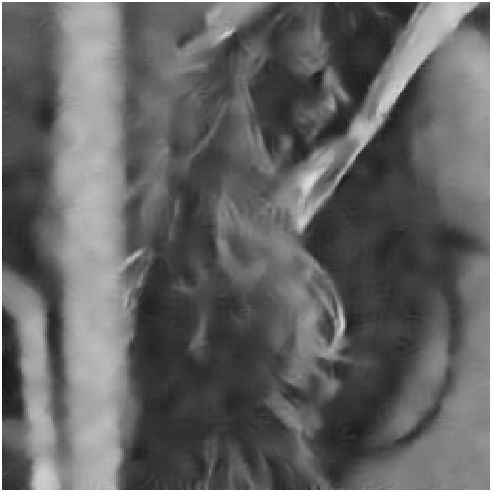}
        \end{subfigure}
        \caption{Close-up denoising results. From the left, NLM, FoE, and ensemble learning}\label{fig:close_up}
\end{figure}

The denoising results of NLM often perform well at recovering sharp edges. In contrast, FoE often produces blurry edges by comparison, especially when the input noisy sigma value is high. However, mottling can sometimes be found on the denoising results of NLM. Ensemble learning is able to preserve the benefits of both techniques, retaining sharp edges with minimal mottling. Therefore, some disadvantages from NLM and FoE could be mitigated by the ensemble learning.

Figure \ref{fig:close_up} shows the close-up denoising results. Figure \ref{fig:close_up} demonstrate the advantages and disadvantages of NLM and FoE. Sharp edges can be found in the result of NLM, but not in the result of FoE. Mottling could be detected in the result of NLM. The ensemble learning model retains sharp good edges but also has less mottling. In other words, the ensemble learning could overcome some disadvantages from NLM and FoE.

\begin{figure}[ht]
        \centering
        \begin{subfigure}[b]{0.45\textwidth}
                \centering
                \includegraphics[width=\textwidth]{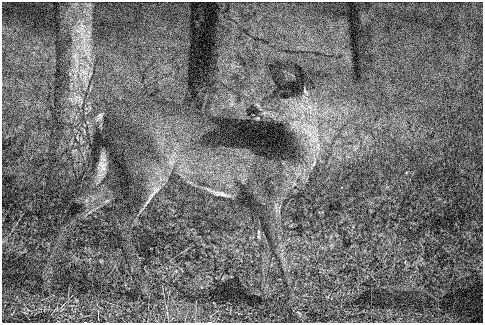}
        \end{subfigure}%
        \quad 
        \begin{subfigure}[b]{0.45\textwidth}
                \centering
                \includegraphics[width=\textwidth]{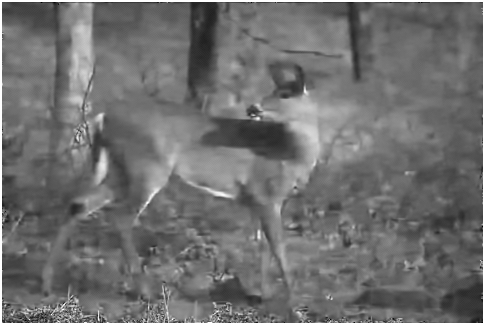}
        \end{subfigure}
        ~ 
        \begin{subfigure}[b]{0.45\textwidth}
                \centering
                \includegraphics[width=\textwidth]{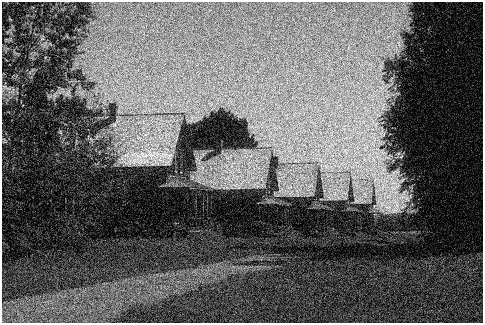}
        \end{subfigure}
        ~ 
        \begin{subfigure}[b]{0.45\textwidth}
                \centering
                \includegraphics[width=\textwidth]{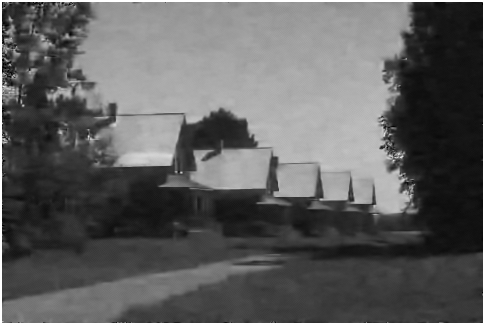}
        \end{subfigure}
        \caption{Other ensemble learning results with $\sigma = 30$ (Left: Noisy image, Right: Denoised image)}\label{fig:other_results}
\end{figure}

\begin{figure}[ht]
        \centering
        \begin{subfigure}[b]{0.47\textwidth}
                \centering
                \includegraphics[width=\textwidth]{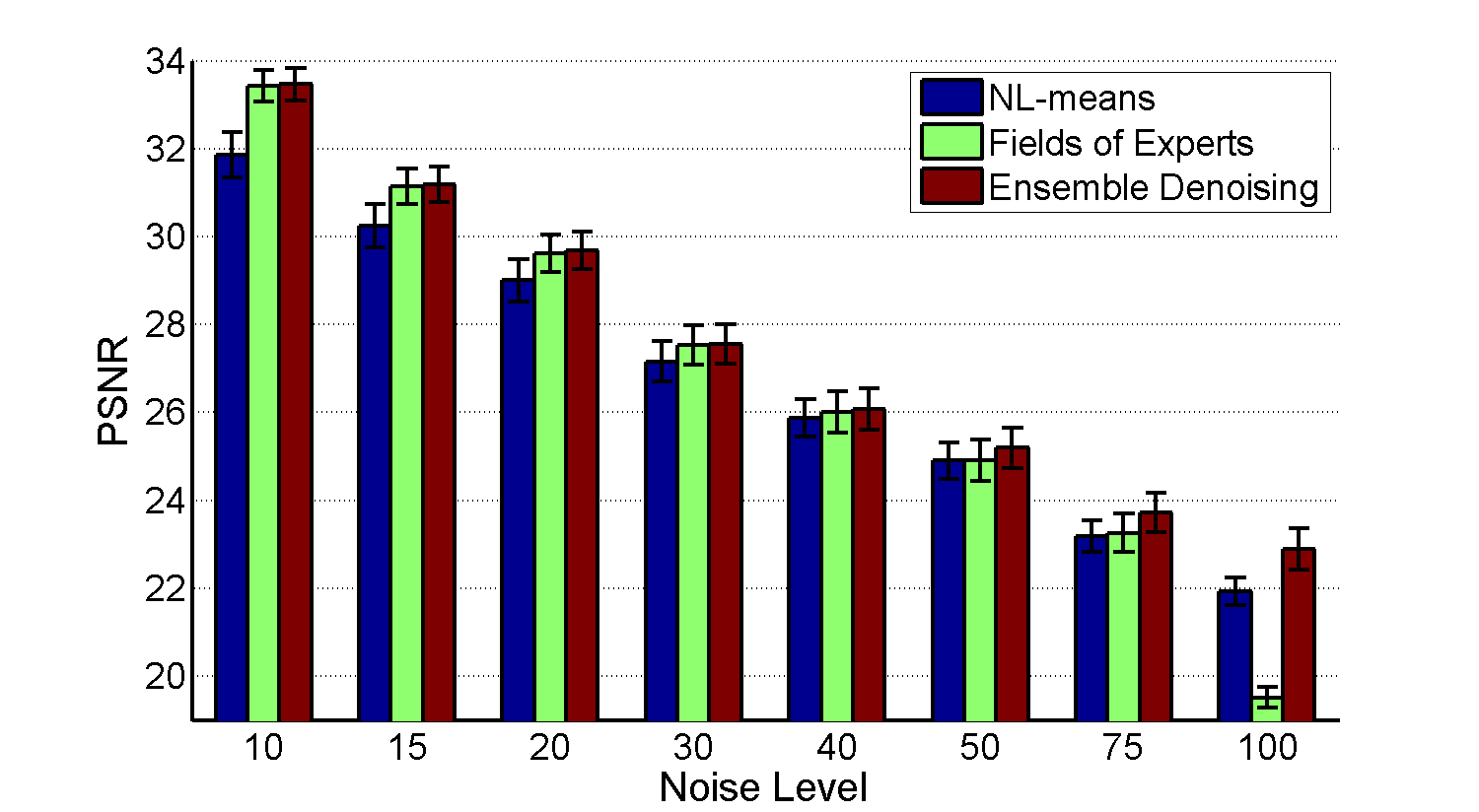}
                \caption{}
                \label{fig:7_a}
        \end{subfigure}%
        \quad 
        \begin{subfigure}[b]{0.47\textwidth}
                \centering
                \includegraphics[width=\textwidth]{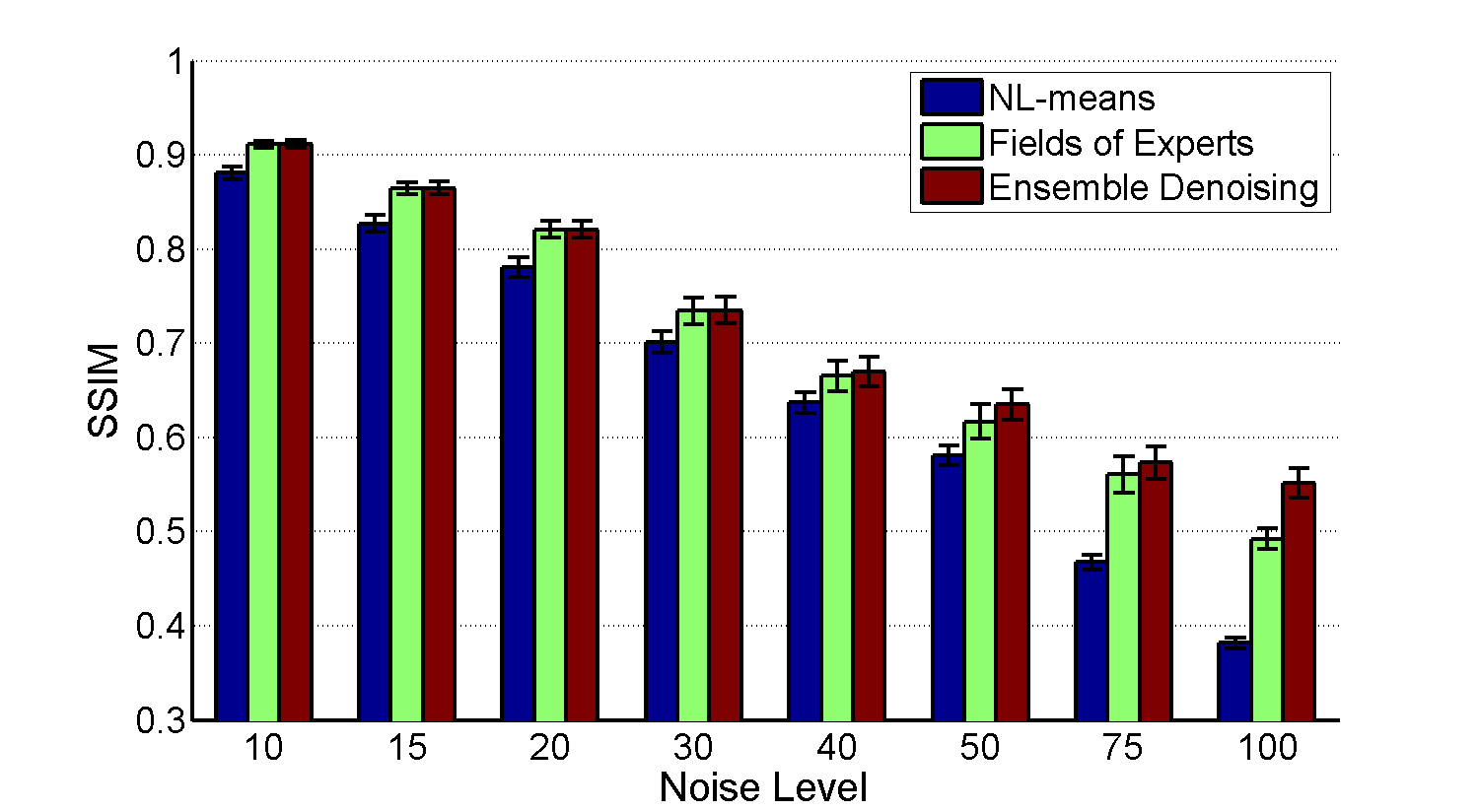}
                \caption{}
                \label{fig:7_b}
        \end{subfigure}
        ~ 
        \begin{subfigure}[b]{0.47\textwidth}
                \centering
                \includegraphics[width=\textwidth]{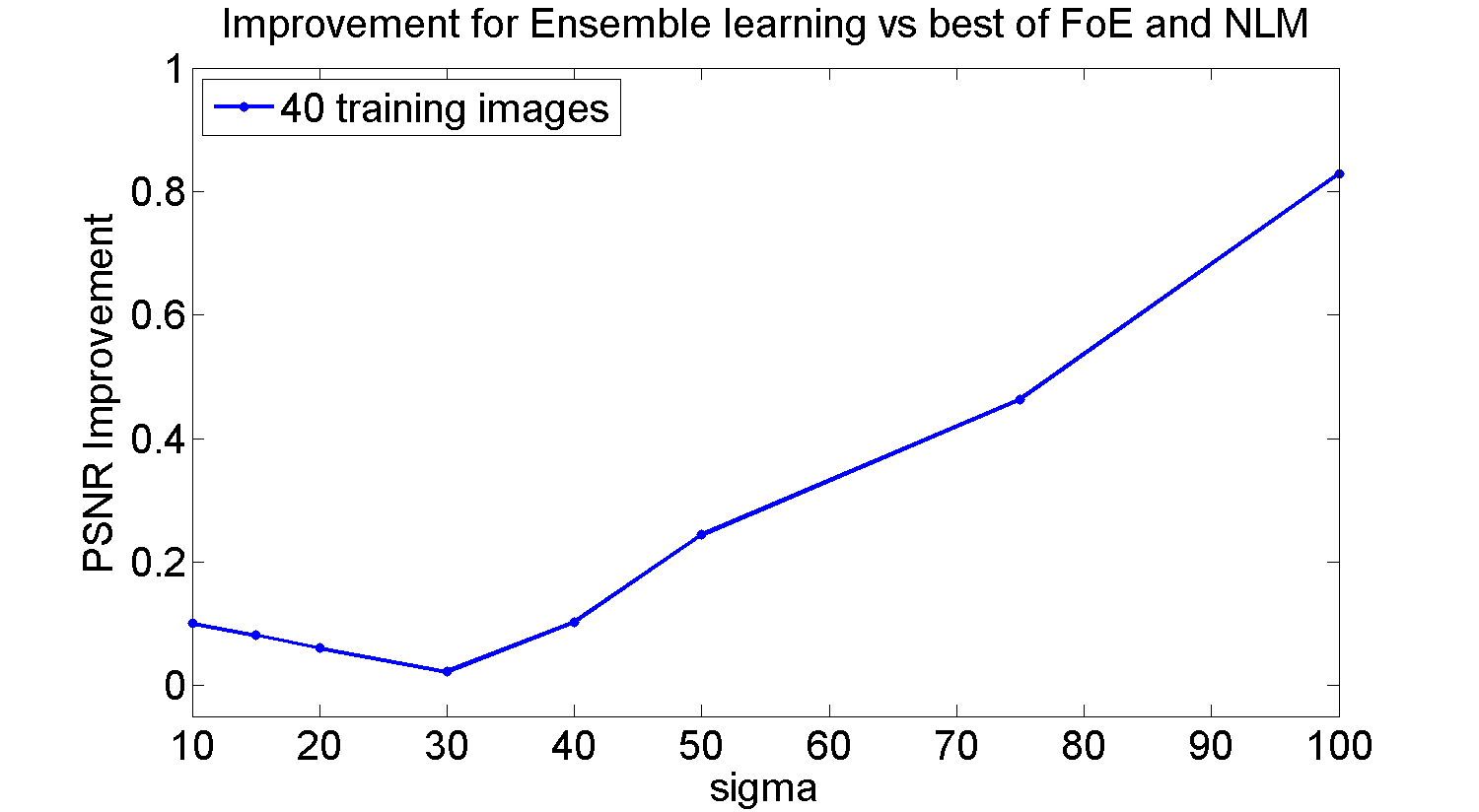}
                \caption{}
                \label{fig:7_c}
        \end{subfigure}
        ~ 
        \begin{subfigure}[b]{0.47\textwidth}
                \centering
                \includegraphics[width=\textwidth]{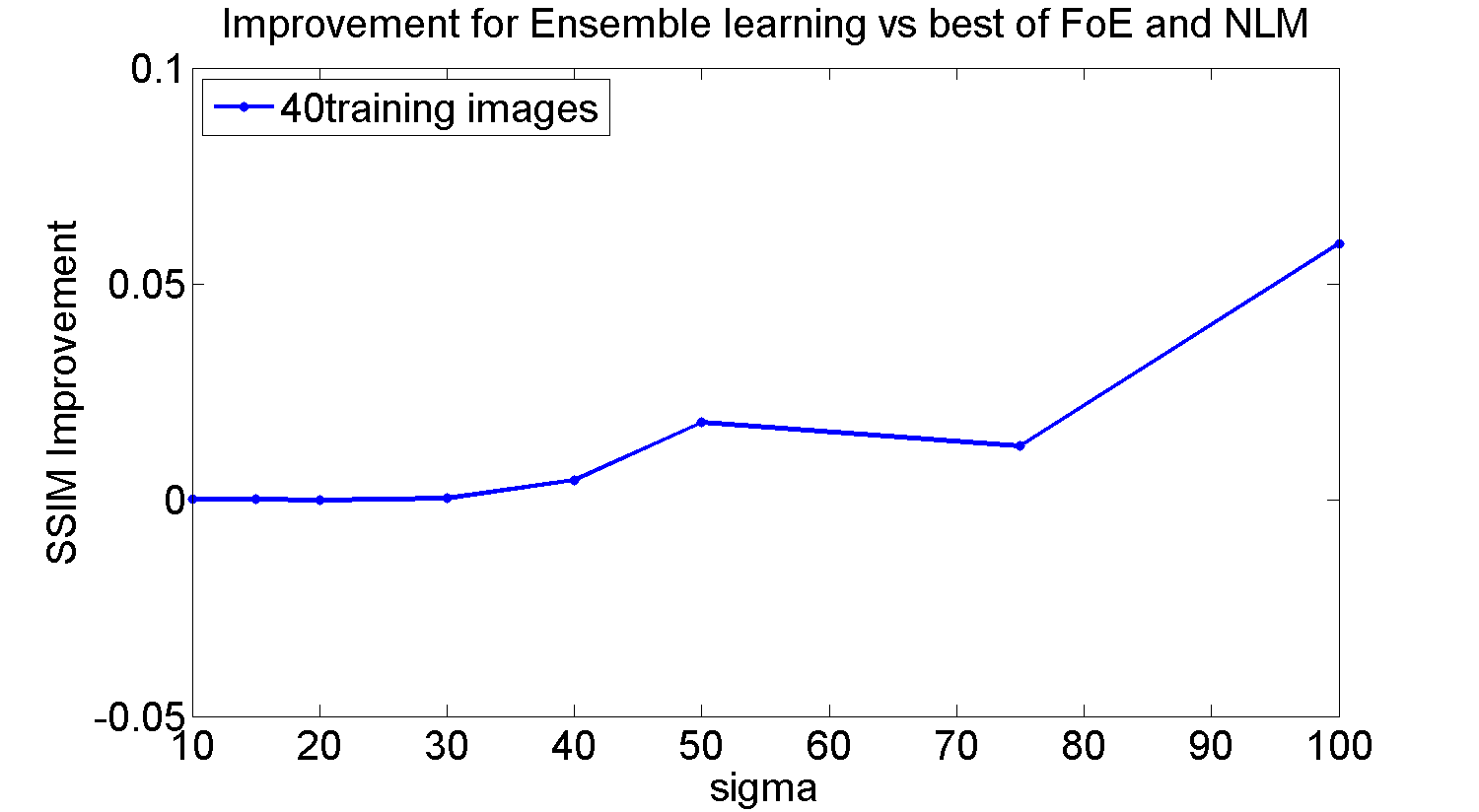}
                \caption{}
                \label{fig:7_d}
        \end{subfigure}
        \caption{Denoising results (a) PSNR in dB. (b) Denoising SSIM results. 
From left to right, Non-local means, Fields of Experts, and Ensemble learning. (c), (d) Improvement of the ensemble learning in PSNR and SSIM. The blue line plots the improvement in performance versus the best performance of either NLM or FoE, whichever method performed better, Values are averaged over all 80 testing images.}\label{fig:bar_graph}
\end{figure}

The second result was done with a subset of 80 images from Berkeley database \cite{martin2001database}. These test images were disjoint from the set of training images. Figure \ref{fig:other_results} shows noisy images and results of ensemble learning with $\sigma = 30$. The result of average PSNR values at different input noise level is displayed in Figure \ref{fig:7_a}. The standard error is shown as the error bar. The ensemble learning outperformed the NLM and FoE in most cases. The ensemble learning showed a better improvement, especially at the high levels of the input noise. When the input noise is 100, the average PSNR results of ensemble learning, NLM, and FoE are 22.75dB, 21.83dB, and 19.47dB, respectively. Figure \ref{fig:7_b} shows the result of average SSIM of 80 testing images. When the input noise is 100, the average SSIM results of ensemble learning, NLM, and FoE are 0.5701, 0.3885, and 0.5062, respectively. Figure \ref{fig:7_c} and \ref{fig:7_d} show an improvement of the ensemble learning versus the best value of FoE and NLM. Ensemble learning shows a consistent improvement in PSNR over the two constituent denoising methods, and improvement increases as the input sigma increases.

\section{Conclusion}
Using a novel Bayesian ensemble learning method, we were able to achieve image denoising that outperformed either of the constituent denoising technique. The results showed that the ensemble learning quantitatively had outperformed the NLM and FoE. The majority of PSNR results for ensemble learning showed an improved output, and its denoising performance augmented the quality of the denoised images when the input noise increased. Although, the PSNR of FoE dropped rapidly as the input noise level increased, ensemble learning was still able to capitalize on the FoE results to produce an improvement over NLM, and the performance of the ensemble method decreased only slowly with sigma, as NLM does. For example, when the input sigma dropped from 75 to 100, the PSNRs of FoE were 23.19dB and 19.47dB, respectively. However, the PSNRs of ensemble learning were 23.62dB and 22.75dB, respectively. Moreover, the ensemble learning method was able to retain specific advantages of each constituent method without losing the benefits of each. The denoising results of FoE sometimes could lead to blur at some edges when the input sigma is high. Mottling sometimes is shown on the denoising images of NLM. However, the ensemble learning is available to remove the mottling from NLM and get better sharpened images at the edges. Therefore, the ensemble learning may have advantages over NLM and FoE.

There remain some features that could be improved to the ensemble learning for our future work. The calculation to get $\sigma_{pseudo}$ and $N_{pseudo}$ is based on the proportional value. In other words, we may need to consider the detailed calculation procedure to get the exact value of the $\sigma_{pseudo}$ and $N_{pseudo}$. Our model for $p(D_{NL}|I)$ is simple, so this could be improved by learning it from natural images. 

In this paper, we focus on testing the probabilistic ensemble learning model on FoE and NLM. These methods were selected in part because they are very different from one another, in both their philosophy of approach as well as their strengths and weaknesses. The ensemble learning method developed here is flexible and can be extended to include a wide variety of denoising methods (such as Gaussian Mixture  or Sparse coding \cite{olshausen1997sparse}, \cite{portilla2003image}). If we add more denoising methods, the formula \ref{eq:11} could be written as follow:
\begin{equation}
\centering
P(I|N, D_{gau}, D_{NL}) = P(N|I, D_{gau}, D_{NL}) \frac{P(I|D_{gau}, D_{NL})}{P(N|D_{gau}, D_{NL})}
\end{equation}•

$P(N|I), P(D_{gau}|I), P(D_{NL}|I)$ are all Gaussian distribution and $P(I)$ is the Fields of Experts model. Therefore, we would be available to calculate $\sigma_{pseudo}$ and $N_{pseudo}$ by using formula \ref{eq:14} \ref{eq:15} \ref{eq:16} \ref{eq:17}. Future work will include further improvements by enhancing these features of the ensemble learning.





\bibliographystyle{model1a-num-names}
\bibliography{paper_Taek.bbl}







\end{document}